\documentclass[runningheads]{llncs}
\usepackage{graphicx}
\usepackage{comment}
\usepackage{amsmath,amssymb} % define this before the line numbering.
\usepackage{color}
\usepackage{multirow}
\usepackage{booktabs}
\usepackage{wrapfig}
\usepackage{tabularx}
\usepackage{caption}
\usepackage{subcaption}
\usepackage[dvipsnames]{xcolor}
\usepackage[pagebackref=true,breaklinks=true,letterpaper=true,colorlinks,bookmarks=false]{hyperref}

\newcommand{\myparagraph}[1]{\vspace{3pt}\noindent{\bf #1}}

\newcolumntype{L}[1]{>{\raggedright\arraybackslash}p{#1}}
\newcolumntype{C}[1]{>{\centering\arraybackslash}p{#1}}
\newcolumntype{R}[1]{>{\raggedleft\arraybackslash}p{#1}}

\begin{document}

% \linenumbers
\pagestyle{headings}
\mainmatter

\title{Synthetic Convolutional Features for \\ Improved Semantic Segmentation} % Replace with your title

\newcommand{\yang}[1]{\textcolor{cyan}{Yang: #1}}
\newcommand{\mario}[1]{\textcolor[rgb]{1,0,1}{Mario: #1}}
\newcommand{\bernt}[1]{\textcolor[rgb]{1,0,0}{#1}}

\titlerunning{Synthetic Convolutional Features for Improved Semantic Segmentation}
\authorrunning{He et al.} 

\author{Yang He\inst{1,2} \and
Bernt Schiele\inst{2} \and Mario Fritz\inst{1}}

\institute{CISPA Helmholtz Center for Information Security \and
Max Planck Institute for Informatics \\\
Saarland Informatics Campus, Germany \\
\email{\{yang.he, fritz\}@cispa.saarland, schiele@mpi-inf.mpg.de}}

\maketitle

\begin{abstract}
Recently, learning-based image synthesis has enabled to generate high resolution images, either applying popular adversarial training or a powerful perceptual loss. However, it remains challenging to successfully leverage synthetic data for improving semantic segmentation with additional synthetic images. Therefore, we suggest to generate intermediate convolutional features and propose the first synthesis approach that is catered to such intermediate convolutional features. This allows us to generate new features from label masks and include them successfully into the training procedure in order to improve the performance of semantic segmentation. 
Experimental results and analysis on two challenging datasets \textit{Cityscapes} and \textit{ADE20K} show that our generated feature improves performance on segmentation tasks.

\end{abstract}

\section{Introduction}
\label{sec:intro}
Semantic image segmentation is a fundamental problem in computer vision, and has many applications in scene understanding, perception, robotics and in the medical area. To achieve robust segmentation performance, models usually are trained with data augmentation like flipping and re-scaling to make full use of expensively annotated data.

Recent work leverages synthetic images as data augmentation for computer vision tasks benefiting from capable graphic engines and development of generative modeling \cite{gan14nips}, e.g. for gaze estimation \cite{cvpr2017GAN_gaze} and hand pose estimation \cite{mueller2018ganerated}, etc. However, using synthesized images for semantic segmentation remains challenging, because of the complexity of scenes and exponential combinations of different elements.
Previous work on semantic segmentation with synthetic data~\cite{hoffman2016fcns,sankaranarayanan2018learning,tsai2018learning,saleh2018effective,wu2018dcan} focuses on domain adaptation problems that aim to reduce the distribution gap between synthetic images and real images, instead of improving a segmentation model trained with fully annotated real data. Besides, even though high resolution realistic generated images \cite{intel_refine:ICCV2017,qi2018semi,wang2018high,park2019SPADE} are able to acquire,
better segmentation results has not been shown by training with those generated images, comparing to training with real images.
When inspecting these generated images visually \cite{wang2018high,qi2018semi}, there are still some 
visual artifacts, which affect the low-level convolutional layers significantly. Learning with those regions, low-level representations are probably degenerated, and thus high-level representations are also hard to effectively build on top of them.
As a result, it becomes hard to train a model with such images, which might lead to decreased segmentation performance.

\begin{figure}[t]
\begin{center}
   \includegraphics[trim=0cm 0cm 0cm 0cm, clip=true,width=0.6\linewidth]{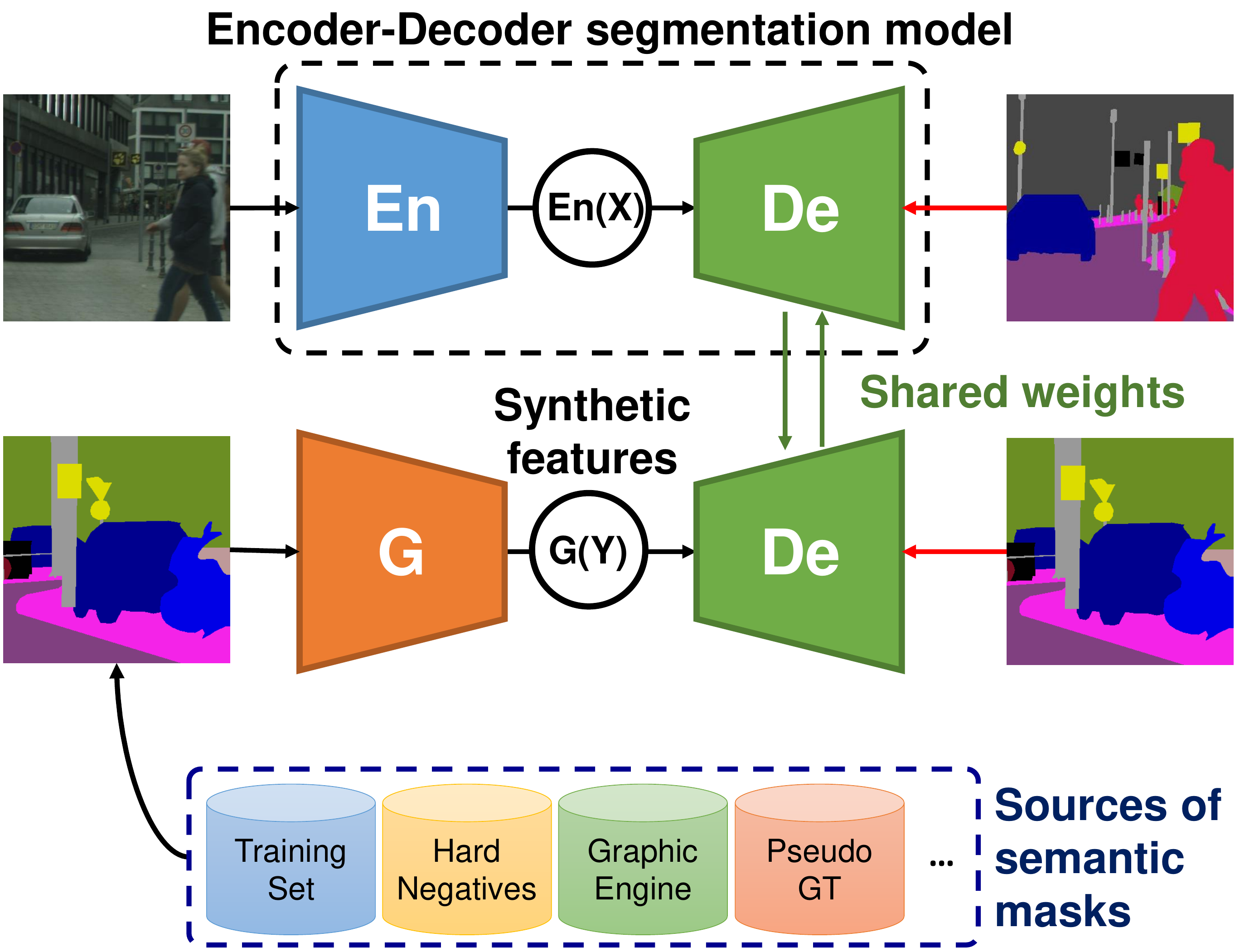}
   \end{center}
   \vspace{-0.2cm}
   \caption{Our pipeline for semantic segmentation. A generator is learned to synthesize convolution features and our semantic segmentation model is trained with synthetic features and real images.}
   \label{fig:overview}
   \vspace{-0.6cm}
\end{figure}

Because of the difficulty of image synthesis, we question if we really need to generate high-quality images for boosting the training of semantic segmentation models.
Instead, we present a feature synthesis-based data augmentation approach for semantic segmentation, as shown in Figure~\ref{fig:overview}. We aim to learn a semantic segmentation model with a mixture of real images and synthetic data from a generator, allowing to sample paired data from semantic layout masks, which assign categories for each pixel. 
Modern semantic image segmentation models built on fully convolutional architectures, as a result, we are able to synthesize convolutional features to approximate the distribution of real features with a similar generation pipeline to images~\cite{zhu2017multimodal}. 
Hence, those synthetic features are able to used as additional training data to improve semantic segmentation. 

The capacity and quality of data is the key to success for data augmentation.
Different to image synthesis, the designed synthetic features have lower spatial dimension but a larger number of channels. 
Consequently, it is hard to directly apply existing image synthesis architectures for the feature generation task, and thus a new effective architecture is needed. 
An ideal feature generator is supposed to meet the following requirements.
(1) It allows us sample multiple diverse features from a semantic mask input, and thus provides us numerous training pairs.
(2) Those synthetic features should follow a similar distribution as extracted features from real images. In other words, the synthetic features are able to be segmented by a trained model, with comparable performance to real features. (3) The final challenge is that raw images contain many detailed information which are compressed in the feature domain. The generator should be powerful enough to represent those important details.
To achieve above goals, we design a generator under the framework of multi-modal translation \cite{zhu2017multimodal} with  a network architecture catered to the convolutional feature synthesis task.

The main contributions of this work are: (1) We propose to synthesize convolution features for data augmentation for semantic image segmentation, leading to improved results.
(2) We present a feature generative model, and analyze its  effectiveness according to a series of ablation studies.
(3) Several techniques are proposed to leverage the synthetic features, including online hard negative mining, generation from additional masks, and label smoothing regularization.

\section{Related Work}
\label{sec:related_work}
\vspace{-0.2cm}
\myparagraph{Generative Adversarial Networks (GAN) and image translation:} GAN \cite{gan14nips} was proposed to capture arbitrary data distribution with learning a discriminator and a generator in an adversarial way.
It was extended to conditional version \cite{mirza2014conditional} that feeds extra information like class labels into a generator, which allows to model the conditional distribution of data and develop many interesting applications. For example, Reed \emph{et al.} \cite{reed2016generative} applied adversarial training to generate different types of flowers from labeled attribute. Odena \emph{et al.} \cite{odena2017conditional} added an auxiliary classifier to the generator that is supposed to recognize the generated data as the input class of generator, and synthesized 1000 ImageNet \cite{deng2009imagenet} classes. 
Particularly, Isola \emph{et al.} \cite{pix2pix2017} proposed the \textit{pix2pix} framework that performs image translation, such as generating colorful shoe images from skeleton images, or generating realistic street scenes from semantic layouts. Besides, Zhu \emph{et al.} \cite{zhu2017unpaired} added a cycle-consistency constraint to achieve unpaired image translation. Recently, Zhu \emph{et al.} \cite{zhu2017multimodal} model the distribution of latent representation under the \textit{pix2pix} framework and generate multiple output images from different modalities, which is called \textit{BicycleGAN}.
Different to \cite{zhu2017multimodal}, we aim to generate dense features, which has totally different dimension in spatial and channel. 

\myparagraph{Semantic segmentation with GAN:} GAN has been applied to semantic segmentation with respective to providing additional loss term \cite{luc2016semantic} or leveraging unlabeled training data \cite{souly2017semi,hung2018adversarial}. Luc \emph{et al.} first applied GAN into semantic segmentation area, which learned a discriminator taking posteriors from a segmentation model as the input, and tried to fool the discriminator with the posteriors. Thus the discriminator provides additional loss term to semantic segmentation model and it is updated within the adversarial training. Besides, people made efforts in leveraging unlabeled data and then lead to semi-supervised learning settings. On one hand, unlabeled data provide a real distribution of natural images for adversarial training, and they might be helpful to successful train a GAN model. One the other hand, the discriminator is able to provide penalty gradients for those unlabeled data, thus it is possible to utilize more data to improve the performance.

\myparagraph{Data augmentation with GAN:} There are several works utilizing generated data with GAN in computer vision tasks \cite{antoniou2017data,frid2018synthetic,xian2018feature_zsl,sixt2016rendergan,peng2018jointly,zheng2017unlabeled,bowles2018gan,cvpr2017GAN_gaze,mueller2018ganerated}.  Xian \emph{et al.} \cite{xian2018feature_zsl} proposed to generate embedding visual features from attributes with GAN for zero-shot learning. They successful trained a classifier with mixing synthetic features for unseen classes and real features of seen classes. As a result, significant improvement was achieved in generalized zero shot image classification problem. 
Besides, Sixt \emph{et al.} \cite{sixt2016rendergan} generated large amounts of realistic labeled images by combining a 3D model. 
Peng \emph{et al.} \cite{peng2018jointly} applied adversarial training to generate many hard occlusion and rotation patterns for augmentation in human pose estimation task.
Zheng \emph{et al.} \cite{zheng2017unlabeled} leveraged large amounts of unlabeled generated images to improve person re-identification task.
GAN was also applied to generate image/label pairs in semantic segmentation.
Bowles \emph{et al.} \cite{bowles2018gan} regard label image as an additional channel, and generate four-channel outputs from a noise, and augment the training set in semantic segmentation.
Finally, \cite{cvpr2017GAN_gaze} and \cite{mueller2018ganerated} address the problem of gaze estimation and hand pose estimation by utilizing the data from rendering system and training a GAN to eliminate the distribution gap between synthetic data and real data.

We highlight that our approach is the first to synthesize dense features, and improve complex task of semantic segmentation by providing synthetic features.
Different to previous work \cite{pix2pix2017,zhu2017multimodal}, features should be successfully used as training data to improve a model, instead of only examples with good visual quality.
Besides, to generate dense features instead of a vector in \cite{xian2018feature_zsl}, we also need more comprehensive architectures.

\section{Semantic Segmentation with Dense Feature Synthesis}
\label{sec:method}
Motivated by the great importance of data augmentation, we depict how to generate dense convolutional features and then leverage the generated features as additional training data to improve semantic segmentation, as illustrated in Fig.~\ref{fig:overview}.
Generation and classification are reverse problems, which translate between images and labels each
other. With a paired training set $\mathcal{T}=\{(\text{X}^i,\text{Y}^i)\}_{i=1}^{n}$, we can learn a segmentation model $\text{Y}^i=S(\text{X}^i)$ as well as an image generator $\text{X}^i=G_{img}(\text{Y}^i)$. Naturally, it is able to train a segmentation model with the augmented dataset $\mathcal{T}\cup \{G_{\text{\it img}}(\text{Y}^i),\text{Y}^i\}_{i=1}^{m}$.

Alternative to generating images, we generate convolutional features for providing more data, and our pipeline is presented in Fig.~\ref{fig:overview}.
Semantic segmentation model $S$ consists of encoder $E$ and decoder $D$, as a result, we can extract features for an image by $E(\text{X})$ and segment the image by $D(E(\text{X}))$. 
Hence, our goal is: (1) to learn a generator $G_{\text{\it feat}}$ which is able to produce realistic features, formally $p(E(\text{X})|\text{Y})\sim p(G_{feat}(\text{Y})|\text{Y})$; (2) to learn the parameters for the segmentation model with the mixture of synthetic and real pairs $\{G_{feat}(\text{Y}^i),\text{Y}^i\}_{i=1}^m \cup\{E(\text{X}^i),\text{Y}^i\}_{i=1}^n$.

In detail, the encoder $E$ is updated with real images only, which captures low-level local features. The decoder $D$ is shared by synthetic and real features, therefore, both steams contribute to the updating of the decoder.
Therefore, the loss function of model training is formulated as
\begin{equation}
\mathcal{L} = \mathbb{E}[-\log D(E(\text{X})|\text{Y})] + \mathbb{E}[-\log D(G_{feat}(\text{Y})|\text{Y})],
\label{eq:loss_overall}
\end{equation}
where the output of $D$ is per class probabilities for each pixel normalized with softmax, and the feature generator $G_{feat}$ is first trained with $\mathcal{T}$ and $E$.

\subsection{Convolutional feature synthesis}
\label{subsec:dense_feature_synthesis}

\begin{figure}[t]
\centering 
\begin{subfigure}{0.45\textwidth} % width of left subfigure
\centering
        \vspace{0.41cm}
		\includegraphics[trim=0cm 10cm 0cm 10cm, clip=true,width=0.98\linewidth]{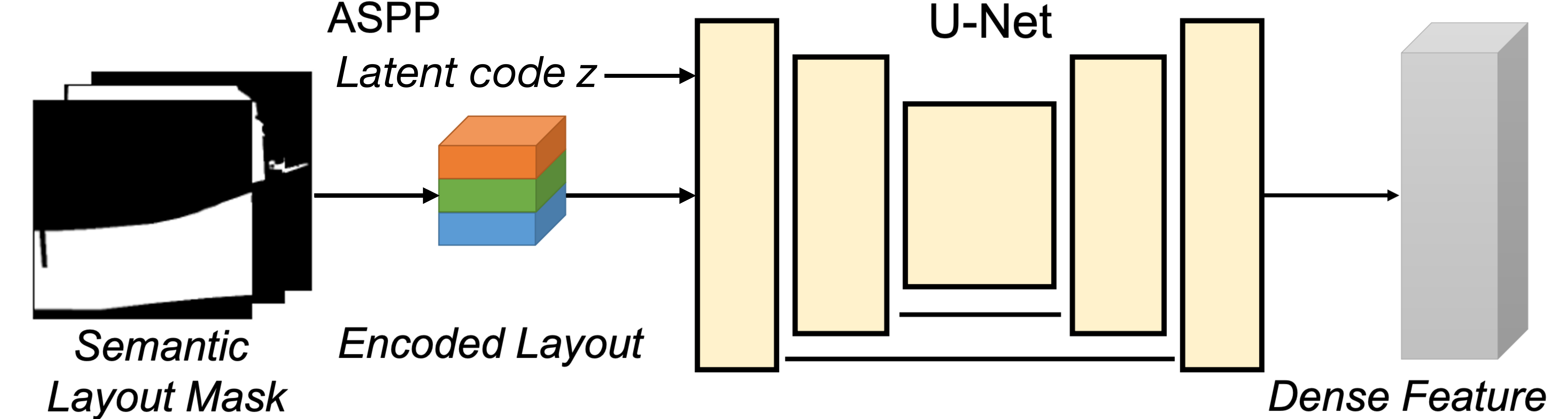}
		\vspace{0.42cm}
		\caption{Generator.} % subcaption
	\end{subfigure}
\begin{subfigure}{0.45\textwidth} % width of left subfigure
\centering
		\includegraphics[trim=0cm 0cm 0cm 0cm, clip=true,width=0.98\linewidth]{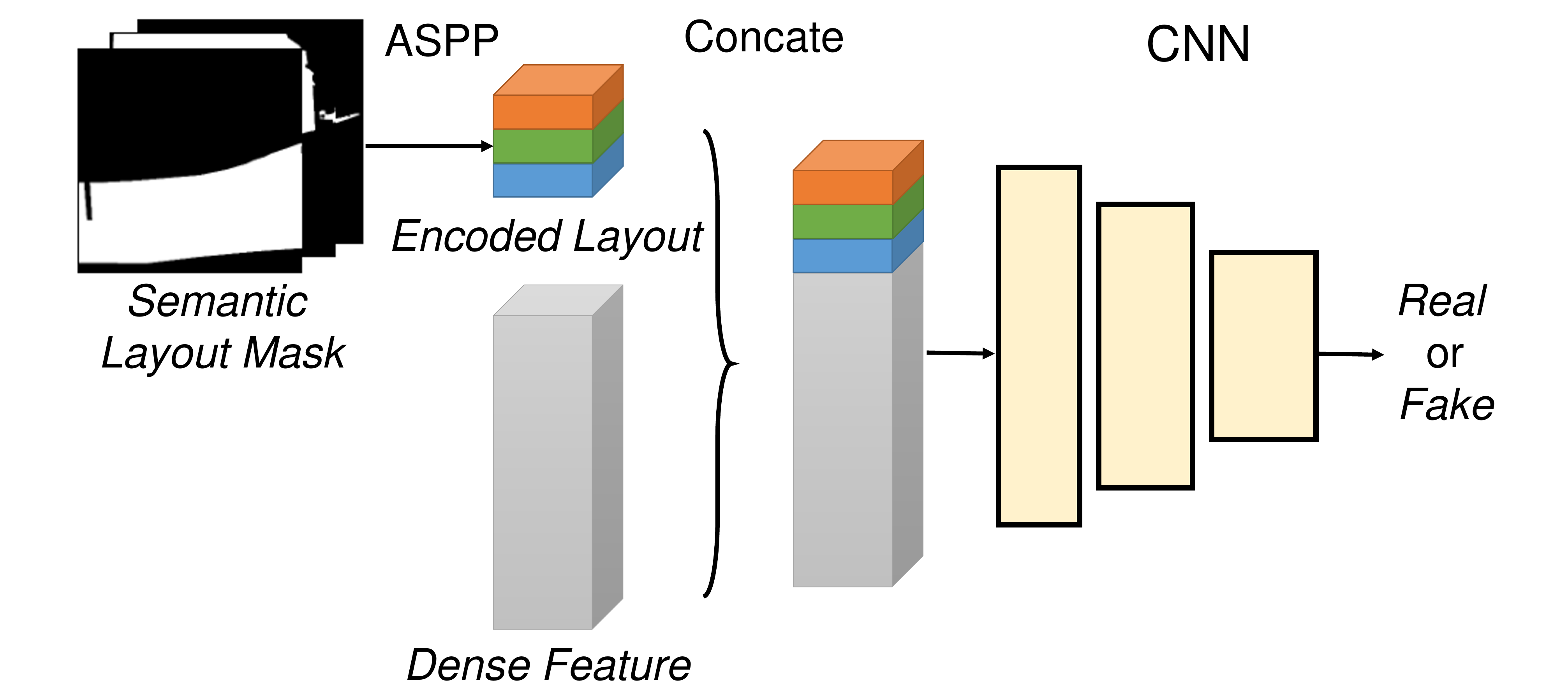}
		\caption{Discriminator.} % subcaption
	\end{subfigure}	
   \caption{The illustration of network architectures in our adversarial feature generator.}
   \vspace{-0.7cm}
\label{fig:generator}
\end{figure}

We present our model for effective feature synthesis, which is crucial for our goal, as discussed in section~\ref{sec:intro}. 
Generating multiple diverse features is challenging, because they encode information of large areas as well as details, which cannot be ignored. 
Also, the synthetic features should follow a similar distribution as extracted real features. We formulate our generator by using the BicycleGAN objective~\cite{zhu2017multimodal}, which is shown to be successful in one-to-many image translation tasks.
In the objective, the generation is driven by a conditional input and a random latent vector, allowing us to sample multiple different examples. Reconstructions on latent vectors and input features guarantee the quality of sampled features. Finally, an adversarial loss helps to generate features with useful details. 

\myparagraph{Architecture.} It turns out that previous synthesis approaches are not directly applicable to feature synthesis, as they emit an output with the same dimension of the input.
Convolutional features are compressed from images with smaller spatial dimension but much larger channel number, encoding location information and many useful details \cite{lrr2016eccv}. Hence, representing such information correctly plays an important role to successful feature synthesis. As shown in Figure~\ref{fig:generator}, our generator takes a high resolution semantic layout mask to produce low resolution feature maps. Our discriminator takes a layout/feature pair as input, to judge if the feature is compatible with the layout or not.

\myparagraph{Preserving resolution by atrous pooling.} Atrous spatial pyramid pooling (ASPP) is an effective module in semantic segmentation used to aggregate multi-scale context information \cite{chen2016deeplab}.
Here, we take advantages of ASPP in multi-scale representation capability, and effectively encode a high resolution semantic layout mask. In our ASPP module, there are three convolution layers with dilation 1,2, and 4, to capture neighboring information and wider context. Stride operation is followed after ASPP, leading to downsampled resolution.
After applying several ASPP modules, the encoded semantic layout reaches to the same spatial dimension as the features. We feed encoded semantic layout and a latent code into a U-Net \cite{unet2015MICCAI} and sample a convolutional feature.
In our discriminator for adversarial training,  
we also apply ASPP module to encode high resolution semantic layout masks. We concatenate the encoded layout and its corresponding real/fake feature together to classify the feature/mask pair as real or fake. 

\subsection{Regularization on synthetic features}
Label smoothing regularization (LSR) has been shown to reduce the influence of noisy labels and improve generalization~\cite{szegedy2016rethinking}. Because not all sampled synthetic features are perfect, we apply LSR on the synthetic features.

In Eq. (\ref{eq:loss_overall}), the per class probabilities for $\text{X}$ are expressively described as
$p_i(k|\text{X}) = \frac{\exp(r_i^k)}{\sum_{k=1}^{K}\exp(r_i^k)}$ for each label $k \in \{1, \cdot\cdot\cdot, K\}$, where $r_i^k$ is the unnormalized log probabilities for $k$-th class, indexing at $i$-th location. Similarly, the per class probabilities for synthetic features $G_{\text{\it feat}}(\text{Y})$ are $p_i(k|G_{\text{\it feat}}(\text{Y})) = \frac{\exp(s_i^k)}{\sum_{k=1}^{K}\exp(s_i^k)}$. The negative log likelihood in Eq. (\ref{eq:loss_overall}) can be rewritten as 
\begin{equation}
\begin{split}
\mathcal{L}  &= \mathbb{E}(-\sum_{i}\log p_i(k|\text{X})q_{real}(k))+\mathbb{E}(-\sum_{i}\log p_i(k|G_{\text{\it feat}}(\text{Y}))q_{syn}(k))
 \end{split}
\label{eq:loss_overall_rewrite}
\end{equation}
with weighting functions $q_{real}(k)$ and $q_{syn}(k)$ for the branches using real images and synthetic features, respectively. For cross entropy loss, it only takes the probability for designed label; for the version with LSR, it takes all the probabilities to compute a loss. They can be formulated in an unified formulation, i.e.,

\begin{equation}
\begin{split}
q_{\epsilon}(k)=
\begin{cases}
    1-\frac{K-1}{K}\epsilon,& k=y\\
    \frac{\epsilon}{K},& k\not=y,
\end{cases}
\end{split}
\label{eq:lsr_weights}
\end{equation}
where $\epsilon$ is a small value in the range of (0,1) for label smoothing regularization. It will become cross entropy when $\epsilon=0$.
As a result, we set $q_{real}=q_{0}$ and $q_{syn}=q_{\epsilon}$ ($\epsilon>0$) in Eq.~(\ref{eq:loss_overall_rewrite}).

\subsection{Online hard negative mining}
\label{subsec:ohnm}
Except generating features by selecting layouts randomly, we can search hard examples during training, which have a large loss value. We do online hard negative mining and feature generation alternatively. We randomly sample some image patches and compute their loss value, the top ranking patches are used to generate the features for the next several training iterations.

\subsection{Additional semantic masks}
Since we generate paired data from semantic layout masks, it is possible to acquire more data than augmenting a training set only, by providing novel mask. For example, in traffic scenarios, the environment is fixed, but everyday road users are different.
It is interesting to know if segmentation model can be further improved by seeing more combination of road users and still objects.
We present more semantic masks from different sources in Table~\ref{table:table_pspnet_validation_cityscapes} of section~\ref{subsec:results_cityscapes}.

\section{Experiments and Analysis}
\label{sec:exp}
\subsection{Experimental settings}
We evaluate our data augmentation scheme using PSPNet \cite{zhao2017pyramid} on the \textit{Cityscapes} \cite{cordts2016cityscapes} and \textit{ADE20K} \cite{zhou2017scene} datasets. \textit{Cityscapes} captures traffic scenes in various cities under different weather and illumination conditions containing 2975 training image pairs with detailed annotations, and 19998 extra images with coarse annotations. Except still frames, it also provides a short video for each frame. \textit{ADE20K} has 20210 training images at different image resolutions including a variety of indoor and outdoor scenes. We evaluate our approach with widely used measurements for semantic segmentation for all the datasets including pixel accuracy (PixelAcc), class accuracy (ClassAcc), mean intersection over union (mIoU) and frequent weighted intersection over union (fwIoU).

\myparagraph{Implementation details}
We implement our generator with the modification of \cite{zhu2017multimodal} using the PyTorch framework.
We implement our segmentation model with synthetic features under the  official PSPNet implementation \cite{zhao2017pyramid}, and apply released ResNet-101 and ResNet-50 based PSPNet as our baselines for \textit{Cityscapes} and \textit{ADE20K}. 

To begin with, we extract 3000 and 40000 patches on the \texttt{conv4\_12} and  \texttt{conv4\_3} layers from official released models to train the generator for \textit{Cityscapes} and \textit{ADE20K}, respectively.
We set 60 and 20 epochs for those datasets. We set three ASPP stages with 96, 192, 384 output feature maps and encode the masks by 8. Besides, we follow the BicycleGAN setup and let the first convolution kernel has 1320 channels.
In addition, we finetune the released model from the baseline with the learned generator which is fixed during the finetuning. All models are learned using SGD with momentum, and the batch size is 16. Initial learning rates are set to $10^-6$ and $10^-7$ for \textit{Cityscapes} and \textit{ADE20K}, and we use the ``poly" learning rate policy where current learning rate is the initial one multiplied by $(1-\frac{iter}{max\_{iter}})^{power}$, and we set power to $0.9$. Momentum and weight decay are set to 0.9 and .0005 respectively. Last, we set $\epsilon$ in Eq.~(\ref{eq:lsr_weights}) for label smoothing regularization is set to 0.0001 and 0.1 for \textit{Cityscapes} and \textit{ADE20K}.

\subsection{Results on Cityscapes}
\label{subsec:results_cityscapes}

\begin{table}[!t]
\centering
\scriptsize
    \begin{minipage}{.5\textwidth}
    \caption{Comparison results on \textit{Cityscapes} validation set. Masks from training (\texttt{T}) and validation (\texttt{V}) set are used to synthesize data.}
    \label{table:table_syn_data_comparison}
    \begin{tabular}{l@{\hspace{0.05cm}}ccccc}
      \toprule
      Models & Mask & PixelAcc  & ClassAcc & mIoU & fwIoU \\
      \cmidrule(lr){1-2}\cmidrule(lr){3-6}
      Baseline &  & 96.34 & 86.34 & 79.73 & 93.15  \\
      \cmidrule(lr){1-2}\cmidrule(lr){3-6}      
      +Img \cite{zhu2017multimodal}  & \texttt{T} & 95.84 & 82.54 & 76.55  & 92.17 \\      
      +Img \cite{wang2018high}  & \texttt{T} & 96.21 & 85.61 & 79.52  & 93.07 \\
      +Img \cite{qi2018semi}  & \texttt{V} & 96.33 & 85.99 & 79.60  & 93.11 \\      
      \cmidrule(lr){1-2}\cmidrule(lr){3-6}      
      Ours & \texttt{T} & 96.40 & 87.29 & 80.30 & 93.27 \\
      Ours & \texttt{T+V} & 96.40 & 87.47 & 80.33  & 93.29 \\
      \bottomrule
    \end{tabular}
    \end{minipage}%
    \qquad
    \begin{minipage}{.45\textwidth}
    \centering
    \caption{Comparison results on \textit{ADE20K} validation set. The top block shows the results of single scale prediction and the bottom is multi-scale prediction.}
    \label{table:table_ade20k_validation}
    \begin{tabular}{l@{\hspace{0.02cm}}ccccc}
      \toprule
      Models &   & PixelAcc  & ClassAcc & mIoU & fwIoU \\
      \cmidrule(lr){1-1}\cmidrule(lr){3-6}
      Baseline &  & 80.04 & 51.75 & 41.68 & 67.46  \\
      Ours & & 80.00 & 53.83 & 42.02  & 68.19 \\   
      \cmidrule(lr){1-1}\cmidrule(lr){3-6}
      Baseline &   & 80.76 & 52.27 & 42.78 & 68.75  \\
      Ours & & 80.81 & 54.70 & 43.35  & 69.17 \\   
      
      \bottomrule
    \end{tabular}
    \end{minipage}
\vspace{-0.5cm}
\end{table}

Table~\ref{table:table_syn_data_comparison} compares models using different sources of synthetic data. During the training of listed models, each batch contains 70\% real images and 30\% synthetic data.
First, we train a segmentation model \cite{zhao2017pyramid}, with using synthetic images from previous state-of-the-art generation approaches \cite{zhu2017multimodal,wang2018high,qi2018semi}.
For \cite{zhu2017multimodal}, we can see the performance is significantly decreased compared to the baseline model \cite{zhao2017pyramid}.
For \cite{wang2018high}, we utilize the training set masks to generate images, which reduces performance across all four metrics comparing the baseline model as well.
In addition, we test another state-of-the-art image generator \cite{qi2018semi}. To know if it is possible to improve semantic segmentation, we even utilize validation set masks. 
Despite providing additional layouts from validation, the performance still decreases at the validation set. Both experiments demonstrate that generated images often do not lead to improved performance.
In contrast, applying our synthetic features successfully improves results by applying different semantic masks.
Particularly, our feature synthesis has the same learning objective to image synthesis \cite{zhu2017multimodal}, however, the performance is able to be improved clearly with training set only as well as additional validation set. The results demonstrate the effectiveness of utilizing our synthetic features for data augmentation in semantic segmentation, while synthetic images are hard to use.

\begin{figure}[!t]
\begin{center}
   \includegraphics[trim=0cm 0cm 0cm 0cm, clip=true,width=0.99\linewidth]{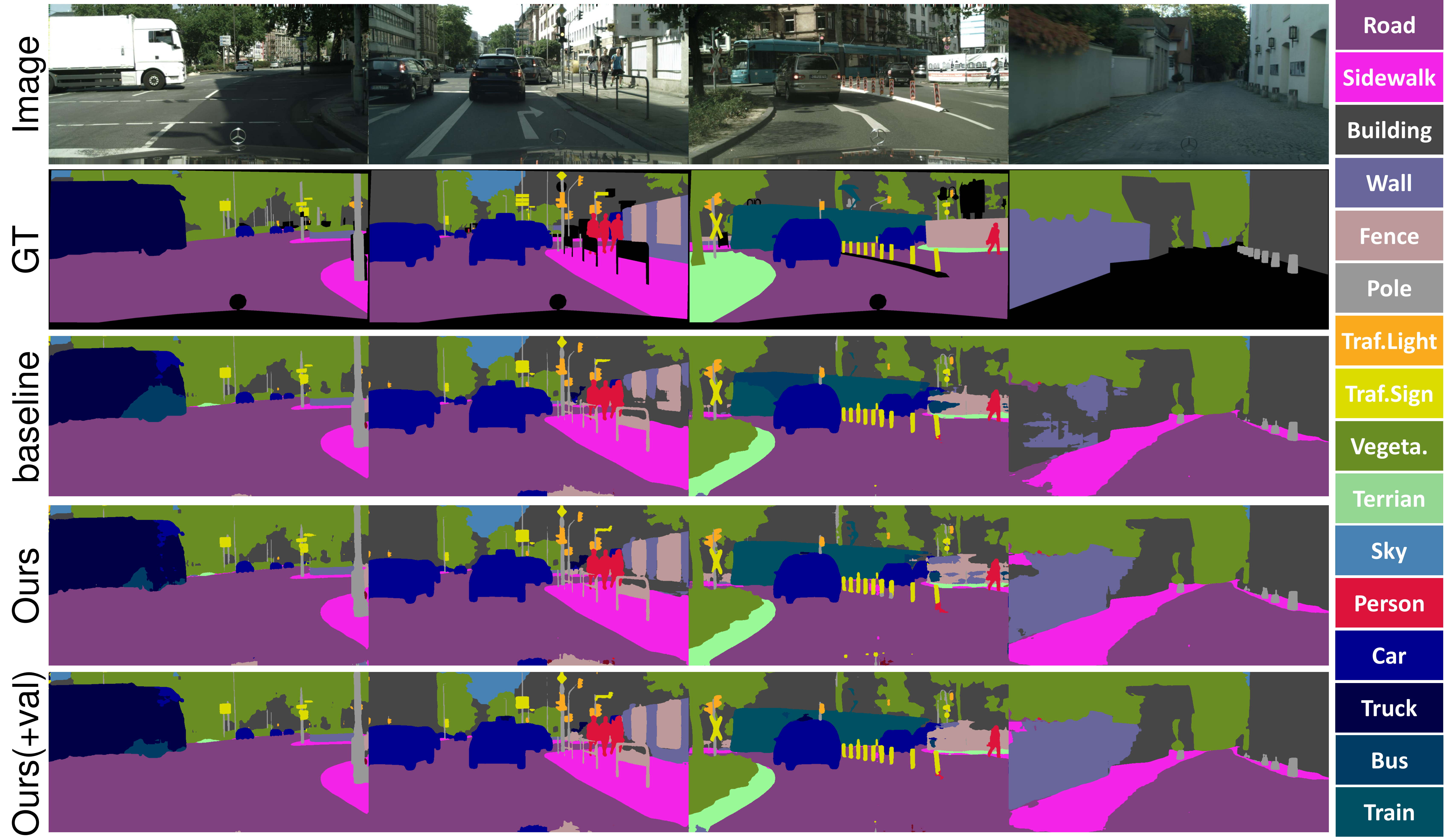}
   \end{center}
   \vspace{-0.4cm}
\caption{Qualitative results on the Cityscapes validation set. We show the augmentation results as well as using additional masks from validation set. Best viewed in color.
}
\label{fig:visualization_comparison_cityscapes}
\end{figure}

In addition to quantitative results, we present some visualization plots compared to PSPNet \cite{zhao2017pyramid} in Fig.~\ref{fig:visualization_comparison_cityscapes}. First, our model predicts more smooth results. 
In the second example, our model successfully recognizes the fence and wall along the street and accurately segment the boundary between them. Besides, for some large regions, i.e., the truck, the train, and the walls in the first, third and fourth examples, our approach achieves clear improvement and segment the whole objects, while baseline predicts unsmooth results.
Particularly, we notice that our model with training masks only is also able to achieve a clear improvement over baseline and very similar results to our model with additional validation masks, which means our data augmentation is very effective.

\begin{figure}[!h]
\begin{center}
   \includegraphics[trim=0cm 0cm 0cm 0cm, clip=true,width=0.95\linewidth]{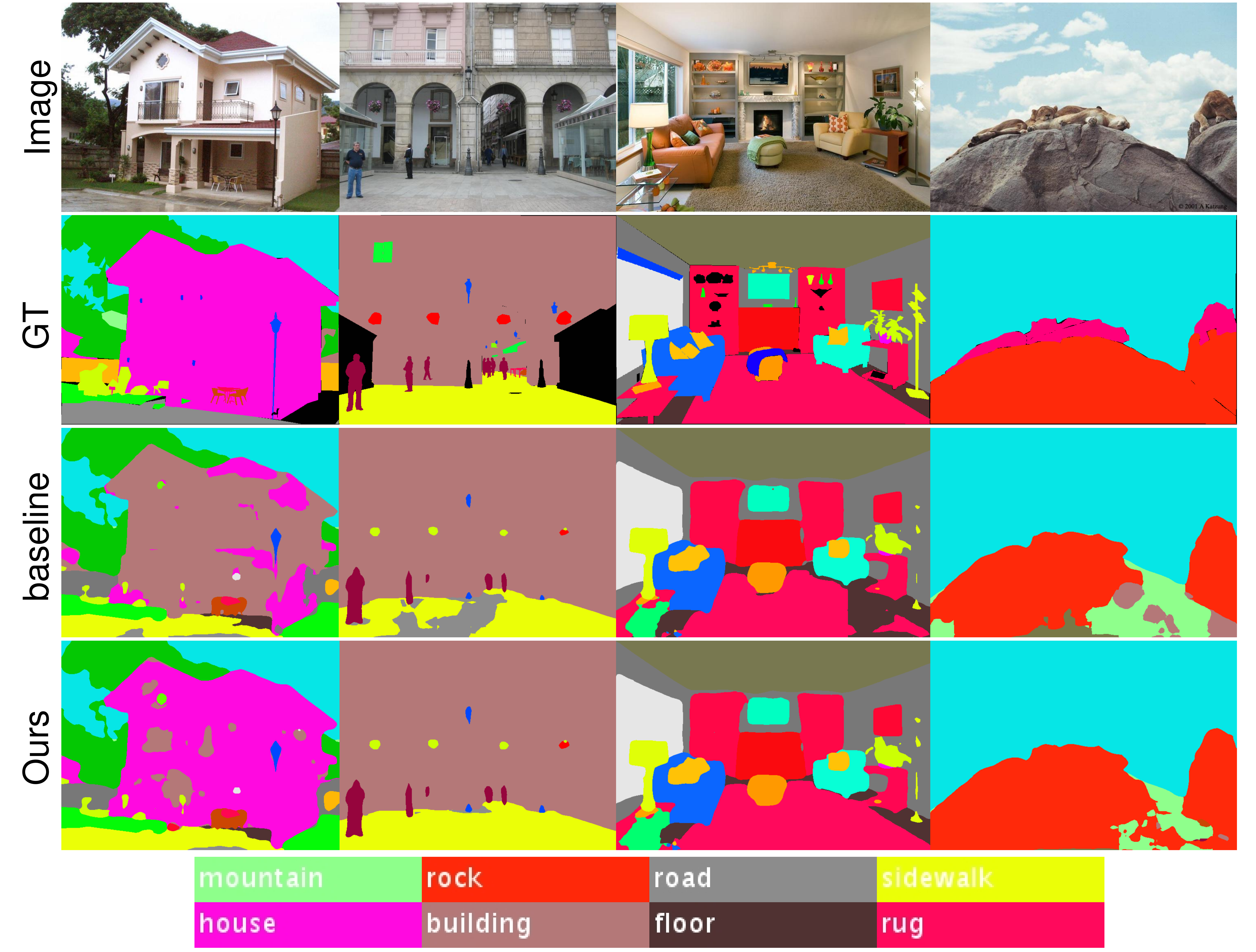}
   \end{center}
\vspace{-0.4cm}
\caption{Qualitative results on the \textit{ADE20K} validation set.  
}
\label{fig:visualization_comparison_ade20k}
\end{figure}

\subsection{Results on ADE20K}
To understand if our approach works well, we also test our approach on \textit{ADE20K} dataset. We present a quantitative comparison in Table \ref{table:table_ade20k_validation} with single-scale prediction as well as multi-scale prediction.
We emphasize that our model achieves 2.08/2.43 and 0.34/0.57 improvements on ClassAcc and mIou for single scale prediction and multiple scale prediction, respectively. Besides, we show several qualitative comparisons in Figure~\ref{fig:visualization_comparison_ade20k}, that our predictions are more smooth and accurate. We are able to distinguish ambiguous classes, such as house/building, mountain/rock. As a result, our model recognizes the entire region for the objects, instead of generating multiple cracked regions.

\clearpage

\subsection{Discussion}
\vspace{-0.6cm}
\begin{figure}[h!]
 \begin{minipage}{0.5\textwidth}
\caption{Per class analysis of ClassAcc and IoU. Yellow indicates more training data, and blue indicates less.
}
\begin{tabular}{ccc}
\rotatebox{90}{\qquad\quad Cityscapes}&
\includegraphics[trim=4.7cm 8.3cm 5.7cm 9cm, clip=true,width=0.49\linewidth]{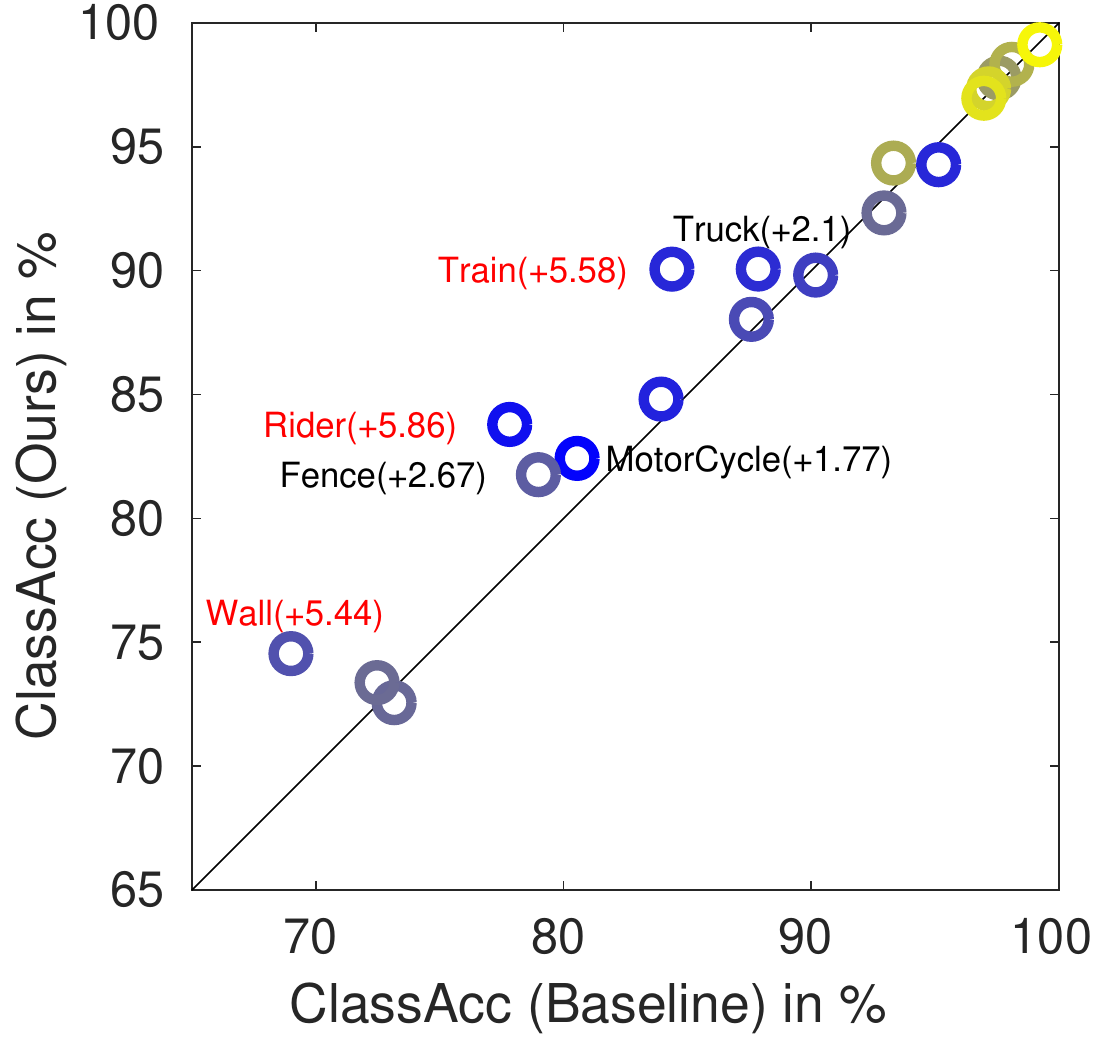}&
\includegraphics[trim=4.7cm 8.3cm 5.7cm 9cm, clip=true,width=0.49\linewidth]{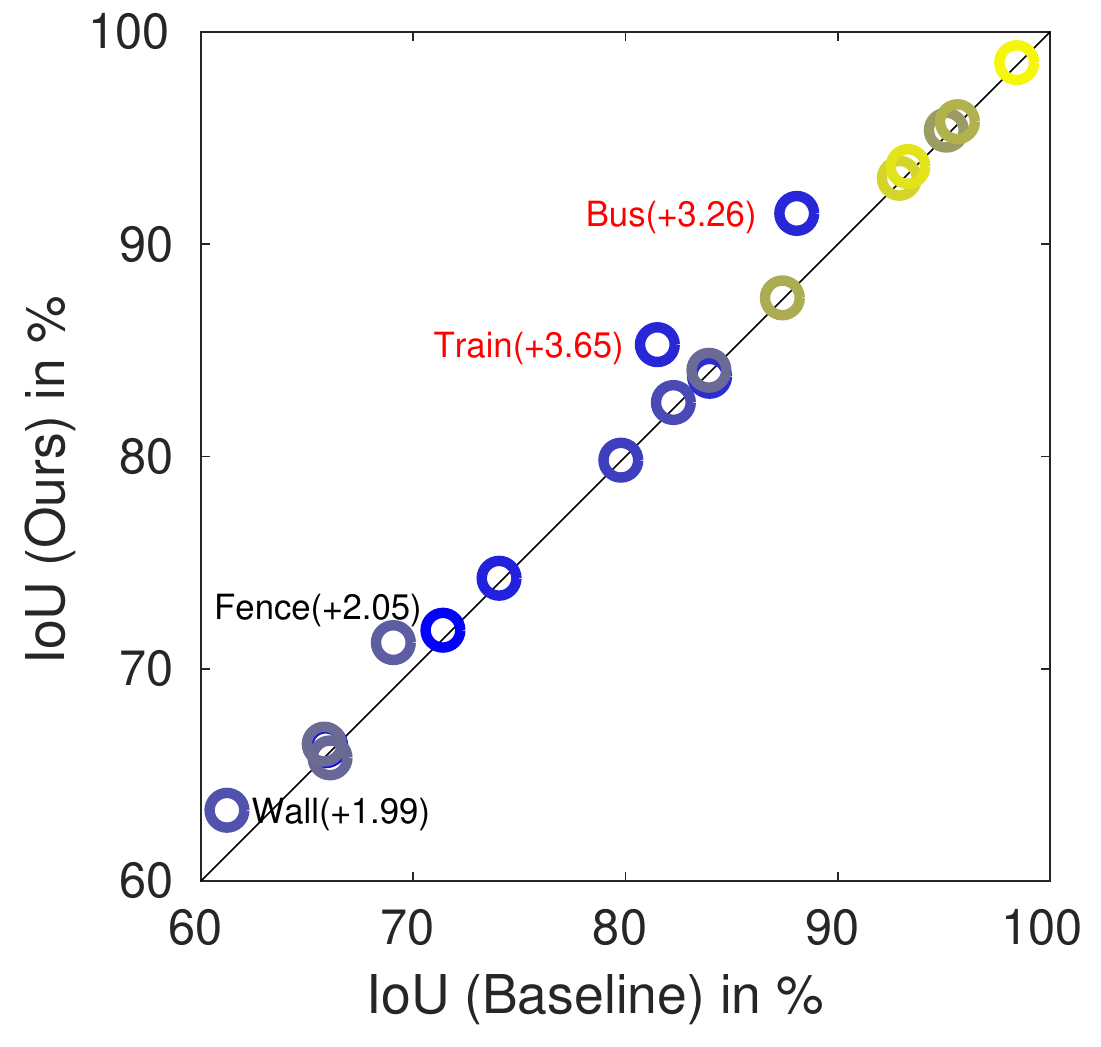}\\
\rotatebox{90}{\qquad\quad ADE20K}&
\includegraphics[trim=4.7cm 8.3cm 5.7cm 9cm, clip=true,width=0.49\linewidth]{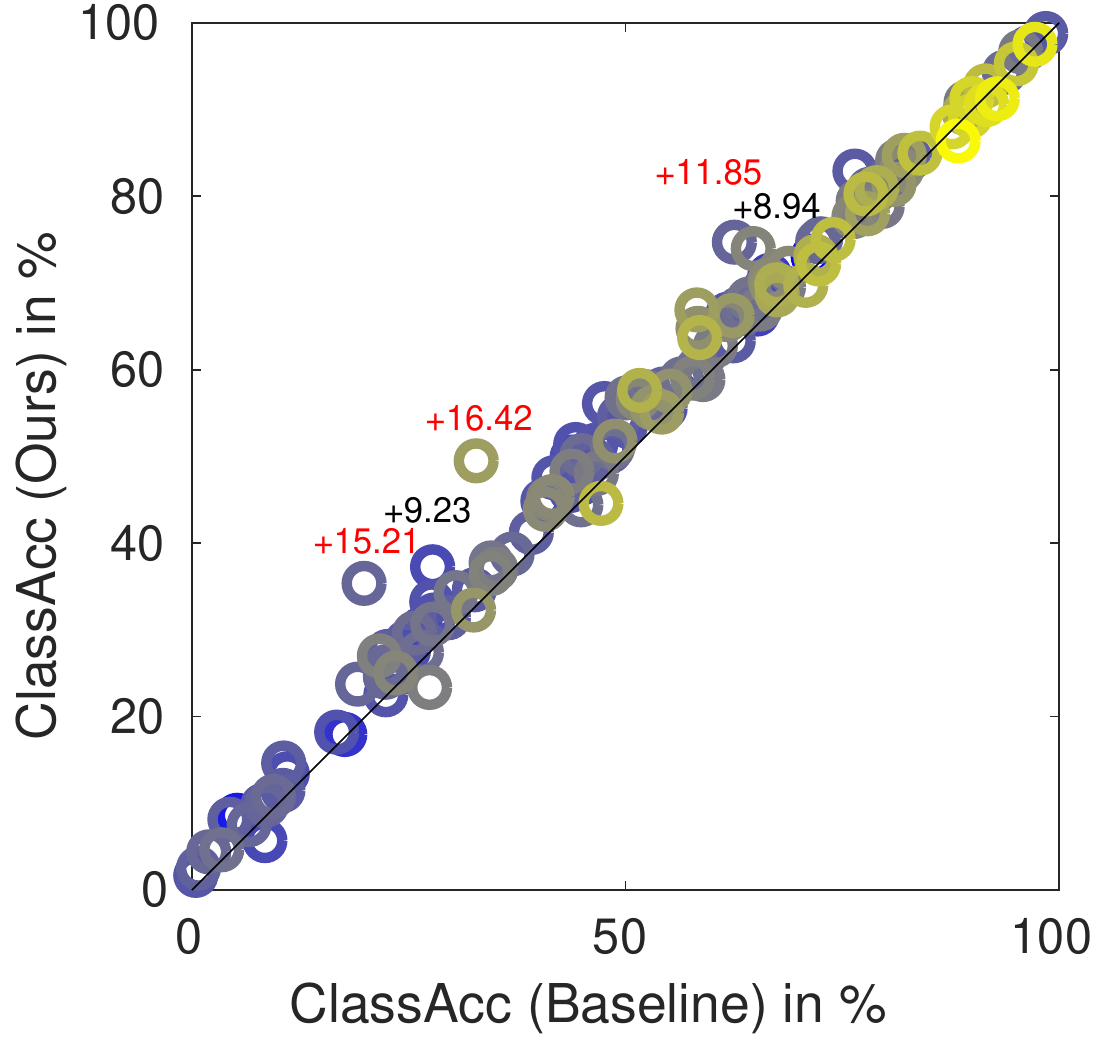}&
\includegraphics[trim=4.7cm 8.3cm 5.7cm 9cm, clip=true,width=0.49\linewidth]{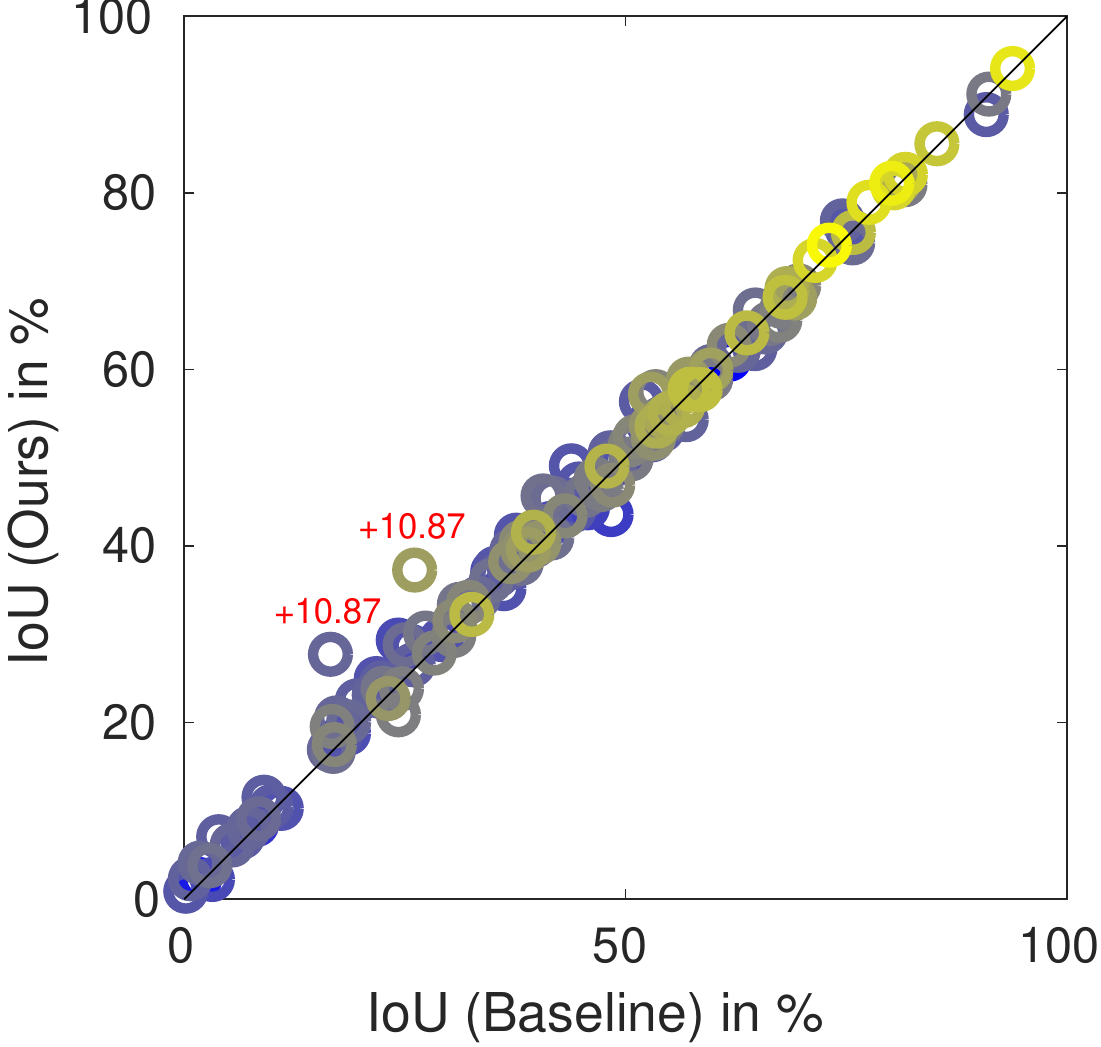}\\
\end{tabular}
\label{fig:per_class_discussion}
 \end{minipage}
 \qquad\quad
 \begin{minipage}{0.43\textwidth}
 \vspace{-0.2cm}
\myparagraph{Per-class analysis} To understand how our method boosts a semantic segmentation model, we analyze the per-class accuracy and
compare the baseline results and our augmentation results. 
From Figure~\ref{fig:per_class_discussion}, we clearly observe substantial improvements for some classes, especially for those with less training data. Besides,there are only a few points significantly falling below the diagonal, i.e., the negative effects are negligible. In \textit{Cityscapes}, Rider, Train and Wall have more than 5\% improvements on ClassAcc; Train and Bus have more than 3\% improvements on IoU. In \textit{ADE20K}, we can see some classes achieved more than 10 points improvement on ClassAcc and mIoU.

 \end{minipage}
\end{figure}

Overall, we show there are 134 classes obtaining improved ClassAcc and 94 classes have improved IoU in all 150 classes.
Particularly, 3 and 2 classes achieve more than 10\% improvements on ClassAcc and IoU, respectively. 
Finally, we also mention that the quantitative comparisons we showed in our main submission and Supplementary Material apparently show the significant improvements.

\myparagraph{Synthetic convolutional features}
To explore the reason for improved results, we analyze the features from different network architectures: (1) baseline architectures in~\cite{zhu2017multimodal}, whose output size is same to input; (2) our architectures described in section \ref{subsec:dense_feature_synthesis}.

\begin{figure*}[!t]
\begin{center}
   \includegraphics[trim=0cm 0cm 0cm 0cm, clip=true,width=0.8\linewidth]{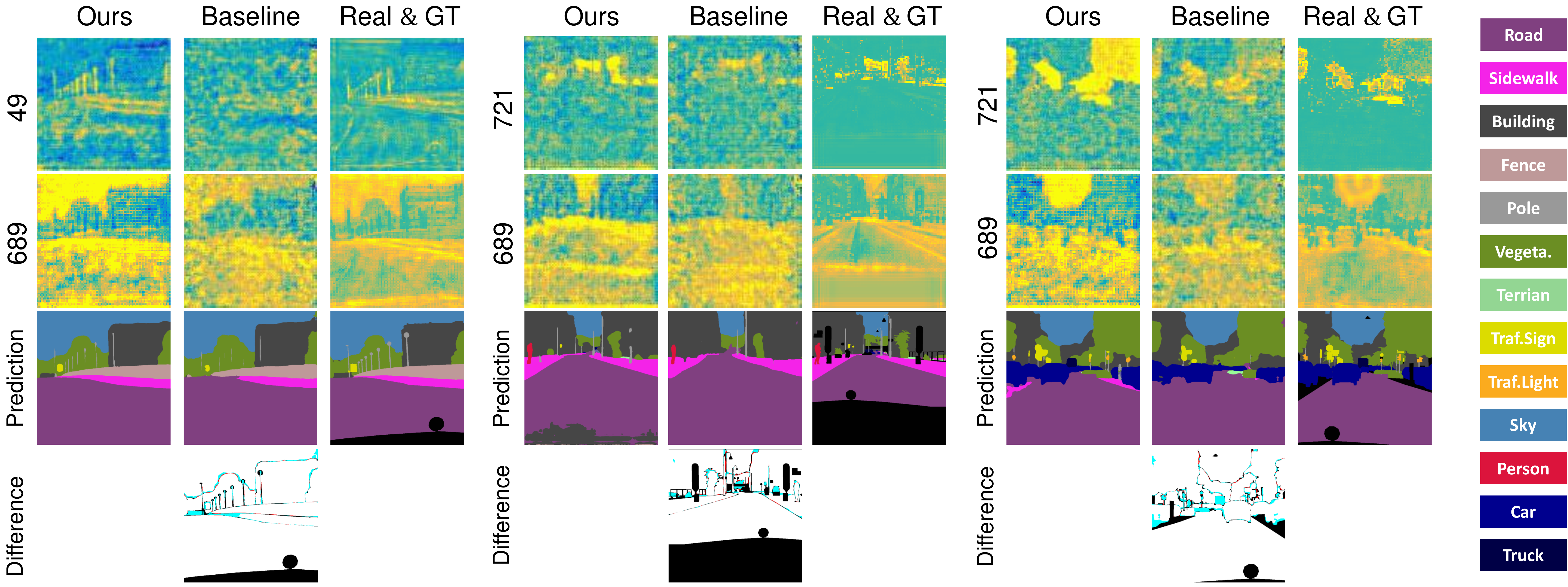}
   \end{center}
     \vspace{-0.1cm}
   \caption{Visualization of synthetic features \texttt{conv4\_12}. In difference maps, cyan color indicates our architecture is better, while red color means baseline is better. Best viewed in color.}
\vspace{-0.6cm}
\label{fig:visualization_features}
\end{figure*}

To begin with, we provide a comparison of PSPNet-scores for those two architectures. We sample 3000 patches from \textit{Cityscapes} training set to compute PixelAcc, ClassAcc, mIoU and fwIoU, as PSPNer-scores.  Second, we train a segmentation model with synthetic features from different architectures. A good generator is supposed to have more close number to real features, and achieves larger improvements. Table~\ref{table:table_aspp} lists those numbers, showing features from our architecture lead to higher PSPNet-scores in four metrics.
For improvements, even baseline has larger ClassAcc, the improvement for PixelAcc, fwIoU is less, and mIoU is quickly decreased.
While using features from our architectures achieves consistent improvements.
As a result, we remark that the effectiveness of our proposed architecture is the key to synthesizing successful features.

To understand more about synthetic features from different architectures, we visualize them as well as the extracted features from real images in Figure \ref{fig:visualization_features}. 
Even though both architectures output activations for different channels at similar locations as real features, our architecture with ASPP is able to produce better details. For example, activations in $49$-th channel are synthesized for the pole class, which is very small, while the baseline cannot do.
In addition, we feed the synthetic features to PSPNet and compare the prediction difference.  Observing all the difference maps, we further conclude that our model with high resolution input is stronger in generating features with accurate boundaries and details.

\begin{table}[t!]
\footnotesize
  \begin{center}
    \caption{Comparison of utilizing different model architectures. }
    \label{table:table_aspp}
    \begin{tabular}{l@{\hspace{0.03cm}}ccccccc}
      \toprule
      Evaluation &  & ASPP & & PixelAcc  & ClassAcc & mIoU & fwIoU \\
      \cmidrule(lr){1-3}\cmidrule(lr){4-8}
      PSPNet-score &  & & & 93.82 & 66.85  &  63.40 & 88.63 \\
      PSPNet-score &  & \checkmark & & 96.44 & 80.05  & 74.33  & 93.44 \\
      Improvement & & &  & +0 & +1.44  & -0.52 & +0.04 \\
      Improvement  & & \checkmark & & +0.06 &  +1.17 &  +0.55 & +0.13 \\
      \bottomrule
    \end{tabular}
  \end{center}
  \vspace{-0.7cm}
\end{table}

Besides, we perform synthesizing different stages of CNN features.
The earliest stage is the image itself, which has shown to be hard to use for data augmentation in Table \ref{table:table_syn_data_comparison}.
To understand the synthesis of good features for semantic segmentation, we extract features (before ReLU layers) at different stages and provide statistics for them from PSPNet, as listed in Table \ref{table:table_pspnet_feature_comparison}. Due to the architecture of PSPNet, the latest stage of features we extract is \texttt{conv4\_23}, which is the input of the pyramid spatial pooling module.

The statistics are collected from 3000 randomly cropped image patches. First, we compute the entropy per class and channel, which is reported the average number in the table. To eliminate the distribution inconsistency for different stages, we normalize all  patches to the same norm ball ($L_2$ norm $100$) for computing the entropy. A good feature that is easier to generate is supposed to have smaller entropy. As we can see, the statistic is consistent to our requirement, that high-level features usually have smaller entropy.

Second, we compute the similarity of activation distributions across classes and channels. We use the intersection of union between two histograms to measure the similarity. As shown in the 4-th rows in Table \ref{table:table_pspnet_feature_comparison}, similarities for higher level features are less, revealing stronger discrimination and ease to generate. Particularly, color pixels have the largest similarity, which further confirms our motivation and is consistent to the difficulty of synthesizing qualified images to augment semantic segmentation, which is consistent to Table \ref{table:table_syn_data_comparison}.

Third, we directly feed the synthetic features to PSPNet to test if they can be recognized by PSPNet. Good synthetic features should be recognized well.
We use mIoU of PSPNet as a measurement, as reported in the 5-th row of the table.
As shown, scores are gradually increased from low-level features to high-level features.

Fourth, we applied different stages of synthetic features as training sets, and report the improvements on ClassAcc and mIou on \textit{Cityscapes} validation set. We observe that improvements on two metrics are obtained by using high level features \texttt{conv4\_6} to \texttt{conv4\_23}. On the contrary, it is hard to achieve clear improvements with low-level synthetic features \texttt{conv1\_3} to \texttt{conv3\_4}. Applying synthetic images leads to the worst performance.
We also observe that the largest improvement comes from \texttt{conv4\_12} instead of \texttt{conv4\_23}. Even \texttt{conv4\_23} has the highest PSPNet-score, it can boost less layers of a segmentation model, comparing to \texttt{conv4\_12}. Finally, we choose to apply synthetic \texttt{conv4\_12} for \textit{Cityscapes} in the rest of this section.

\begin{table*}[t]
\footnotesize	
\begin{center}
    \caption{Statistics on different stages of features. The PSPNet-score for the real is 86.15.}
    \label{table:table_pspnet_feature_comparison}
    \begin{tabular}{ccccccccc}
      \toprule
        & \texttt{Images} & \texttt{conv1\_3} & \texttt{conv2\_3} & \texttt{conv3\_4} & \texttt{conv4\_6} & \texttt{conv4\_12} & \texttt{conv4\_18} & \texttt{conv4\_23} \\
       \cmidrule(lr){1-9}  
       Channels   & 3 & 128 & 256 & 512  & 1024 & 1024 & 1024 & 1024 \\
       Resolution & 1 & 1/2 & 1/4 & 1/8  & 1/8 & 1/8 & 1/8 & 1/8 \\
       Entropy  & 4.8290 & 3.0987 & 3.1584 & 4.0767  & 3.3799   & 3.3659 & 3.2535 & 3.4065 \\
       mIoU of hist.  & 0.5596 & 0.5257 & 0.3171 & 0.3431  &  0.4295  &  0.4087 & 0.3651 & 0.2802 \\
       PSPNet-score  & 2.22 & 9.76 & 26.97 & 56.77 & 70.66  & 74.33 & 61.54 & 83.78  \\  
       $\Delta$ClassAcc  & -3.80 & +0.02 & +0.65 & +0.81 & +0.99   & +1.17 & +1.75 & +0.08  \\      
       $\Delta$mIoU  & -3.18 & -1.09 & -0.37 & -0.1 &  +0.11 & +0.55  & +0.13 & +0.05  \\
      \bottomrule
    \end{tabular}
  \end{center}
  \vspace{-0.8cm}
\end{table*}

\subsection{Additional semantic layout masks}
Since our approach generates a paired data from a semantic mask, we are not only able to augment a training set, but also to leverage new masks to generate more novel examples. To know if more masks are beneficial to boost the performance,
we seek for additional masks as listed in the following:
\begin{itemize}
\item Validation GT masks. It is very easy to acquire high quality masks, which has a similar distribution to the training set. It is used to test if providing more masks with the same distribution is helpful.
\item Rendering system. It provides us large amounts of data at very low cost. In our experiments, we apply recent released \textit{Synscapes} dataset \cite{wrenninge2018synscapes}. Finally, it provides us 25000 extra semantic masks.
\item Pseudo GT masks. We regard prediction from unlabeled images as pseudo ground truth. Although the prediction is not perfect, we generate paired data from a mask, as a result, it still generates features aligned with the input mask. To alleviate the unsmooth prediction, we only save the prediction with posterior larger than 0.7, and complete the holes with nearest neighbor interpolation. To provide novel scenes, we leverage the unlabeled video frames and coarse annotation frames in \textit{Cityscapes}, leading to 29823 extra masks.
\end{itemize}

\begin{table}[t!]
\footnotesize
  \begin{center}
    \caption{Comparison of PSPNet and our approach on Citycapes validation set. The top block is single scale prediction and the bottom is multi-scale prediction.}
    \label{table:table_pspnet_validation_cityscapes}
    \begin{tabular}{l@{\hspace{0.02cm}}ccccc}
      \toprule
      Models  & Additional Mask & PixelAcc  & ClassAcc & mIoU & fwIoU \\
      \cmidrule(lr){1-2}\cmidrule(lr){3-6}
      Baseline  & -- & 96.34 & 86.34 & 79.73 & 93.15  \\
      Ours  & \texttt{No} & 96.40 & 87.29 & 80.30 & 93.27  \\   
      Ours  & \texttt{Val} & 96.40 & 87.47 & 80.33  & 93.29   \\    
      Ours  & \texttt{Synscapes}& 96.39  & 87.55 & 80.03 & 93.27  \\    
      Ours  & \texttt{Pseudo GT} & 96.40 & 87.49 & 80.31 & 93.28 \\       
      \cmidrule(lr){1-2}\cmidrule(lr){3-6}      
      Baseline  & -- & 96.59 & 87.23 & 80.89 & 93.58  \\
      Ours  &  \texttt{No}  & 96.64 & 88.16 & 81.48 & 93.69 \\     
      Ours  & \texttt{Val}  & 96.64 & 88.46 & 81.52 & 93.71 \\    
      Ours  & \texttt{Synscapes} & 96.64 & 88.43 & 81.34 & 93.70  \\    
      Ours  & \texttt{Pseudo GT} & 96.64 & 88.34 & 81.51 & 93.70 \\ 
      \bottomrule
    \end{tabular}
  \end{center}
  \vspace{-0.6cm}
\end{table}

Table \ref{table:table_pspnet_validation_cityscapes} presents the performance with single scale predicton and multi-scale prediction. 
To have a fair comparison, we set the ratio between training masks and additional masks for all the additional mask choices. That is $3:1$, in each sampled batch for synthesizing. Besides, the ratio between real images and synthetic data follows previous experiments $7:3$, and online hard negative mining is applied for various versions of our models.

We observe that after adding validation mask, the performance is further improved than pure augmentation (second row). Besides, applying pseudo GT also achieves better performance than training set only. Interestingly, when we apply the semantic masks from \textit{Synscapes} \cite{wrenninge2018synscapes}, the improvement is less, the reason is the distribution gap of layouts between \textit{Cityscapes} and \textit{Synscapes}. As a result, to leverage synthetic data from the game engine better, not only similar appearance, but also similar semantic layouts are needed, such as shapes, ratios between different objects, etc.

\subsection{Ablation study}
\begin{figure}[!t]
\begin{center}
   \includegraphics[trim=3.0cm 8.5cm 3.5cm 8.0cm, clip=true,width=0.35\linewidth]{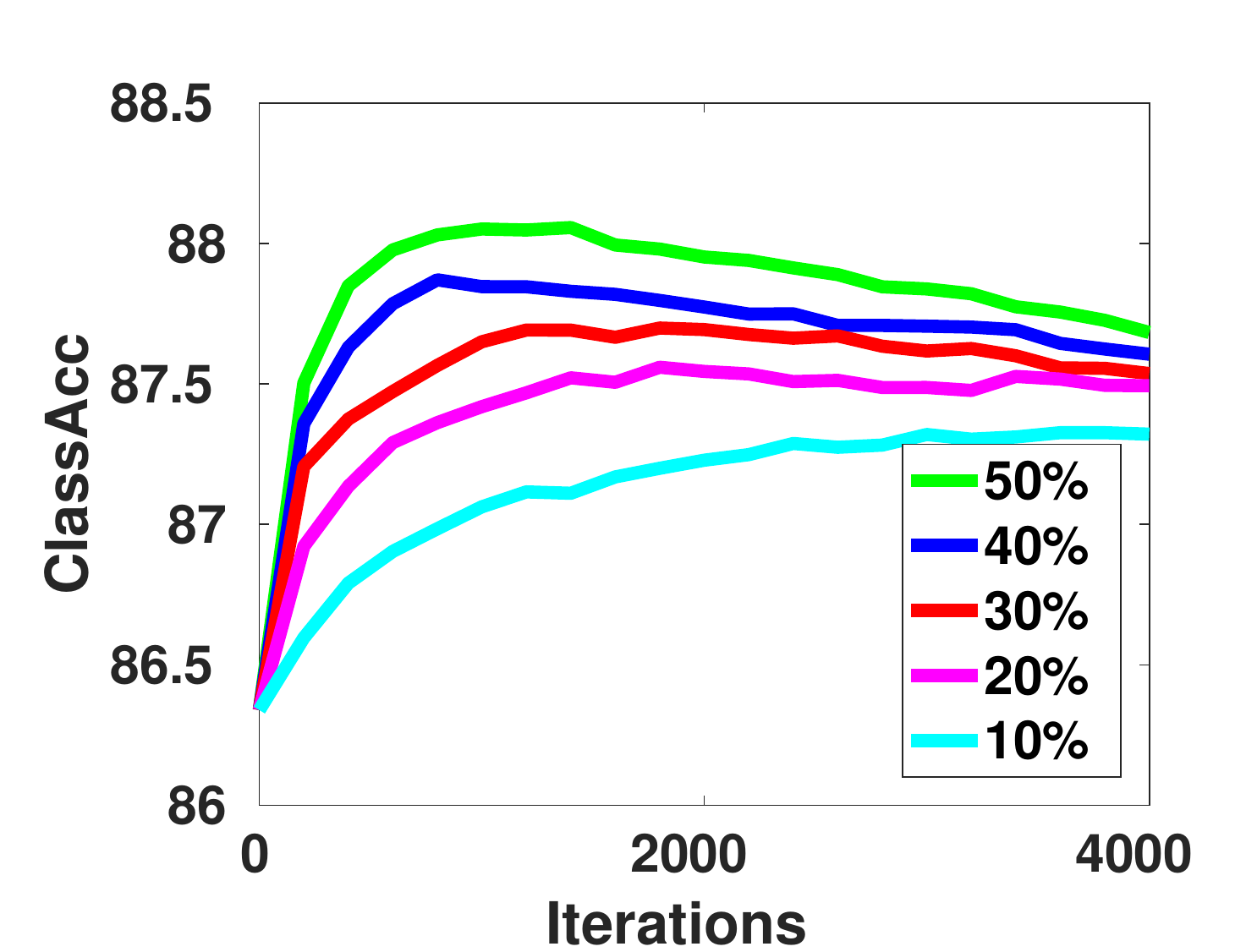}
   \includegraphics[trim=3.0cm 8.5cm 3.5cm 8.0cm, clip=true,width=0.35\linewidth]{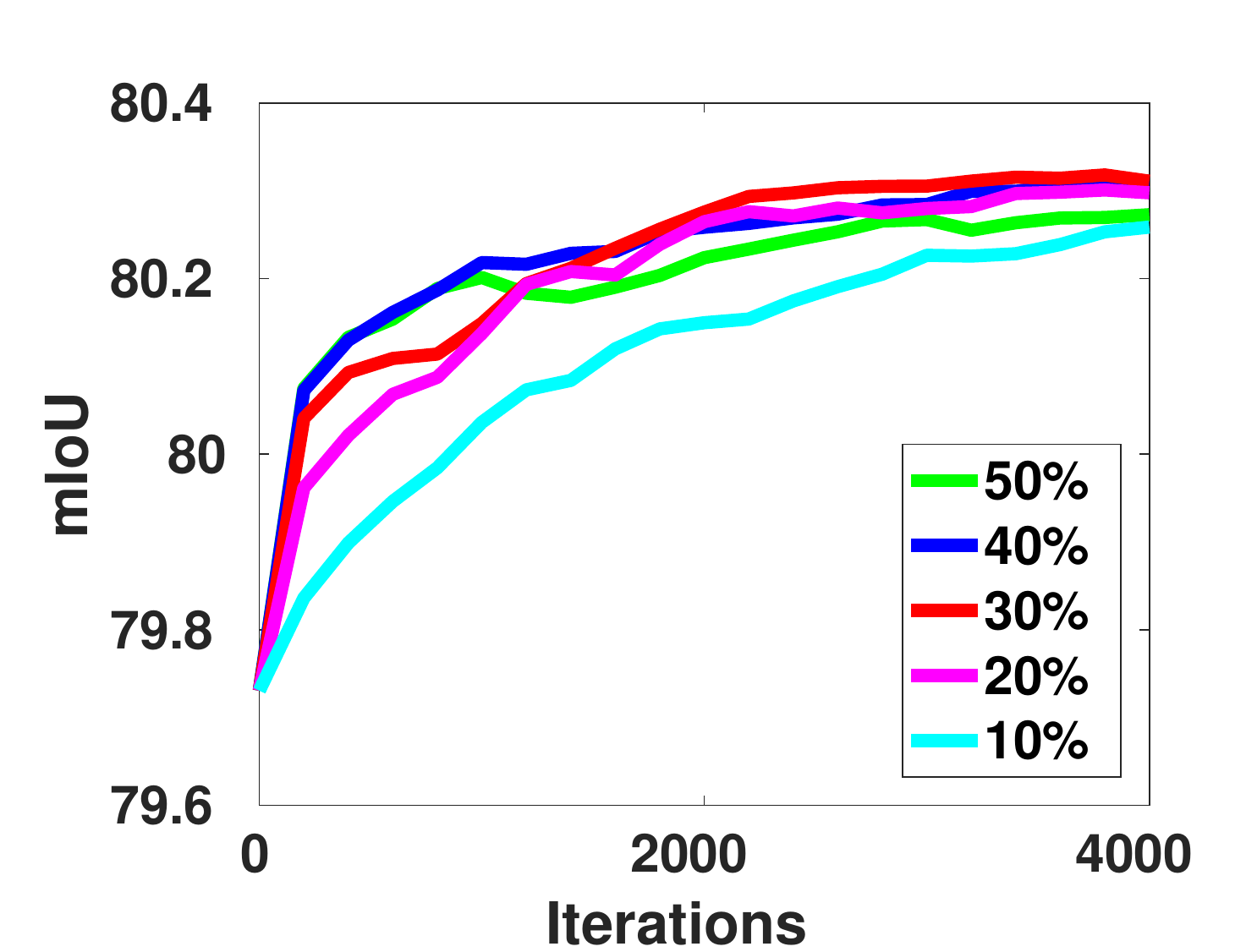}
   \end{center}
   \vspace{-0.4cm}
   \caption{ClassAcc and mIoU on \textit{Cityscapes} at varying training iterations w.r.t. different percentages of our synthetic features in a training batch .}
   \vspace{-0.4cm}
\label{fig:pspnet_cityscapes_percentage}
\end{figure}

\myparagraph{Ratio between real and synthetic data}
In Fig. \ref{fig:pspnet_cityscapes_percentage}, we provide a study on using different percentages of synthetic features in a batch. First, it clearly shows that incorporating our synthetic features with different percentages brings  improvements. Besides, we observe that incorporating more synthetic features will gain more in ClassAcc, which further shows the effectiveness of our approach. On the other hand, better fitting on synthetic data might lead to less improvement on other metrics, as a result, we mix 70$\%$ real images and 30$\%$ synthetic features in each batch for the rest of the experiments.

\begin{figure}[!t]
\begin{center}
   \includegraphics[trim=3.5cm 8.6cm 3.75cm 8.2cm, clip=true,width=0.35\linewidth]{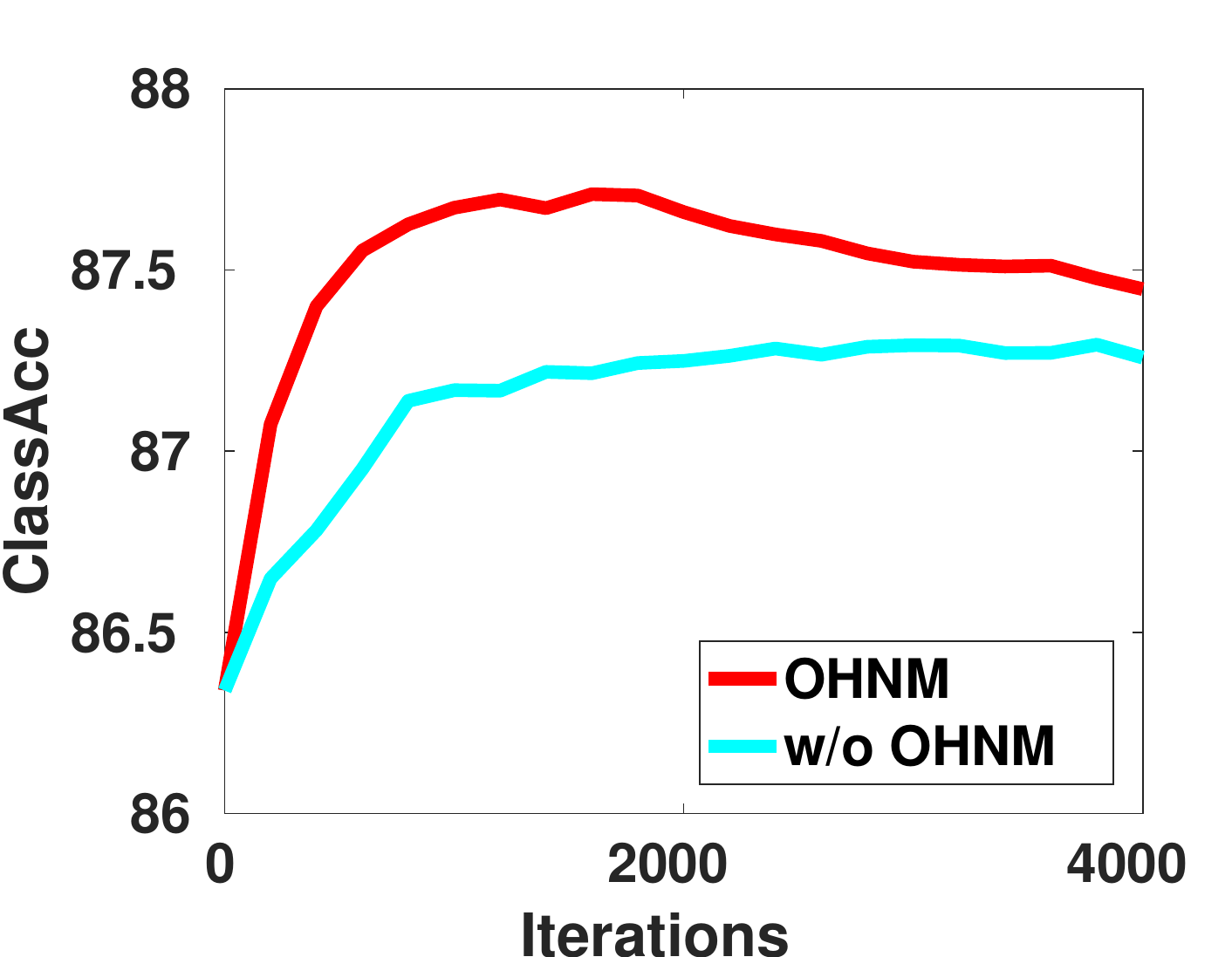}
   \includegraphics[trim=3.5cm 8.6cm 3.75cm 8.2cm, clip=true,width=0.35\linewidth]{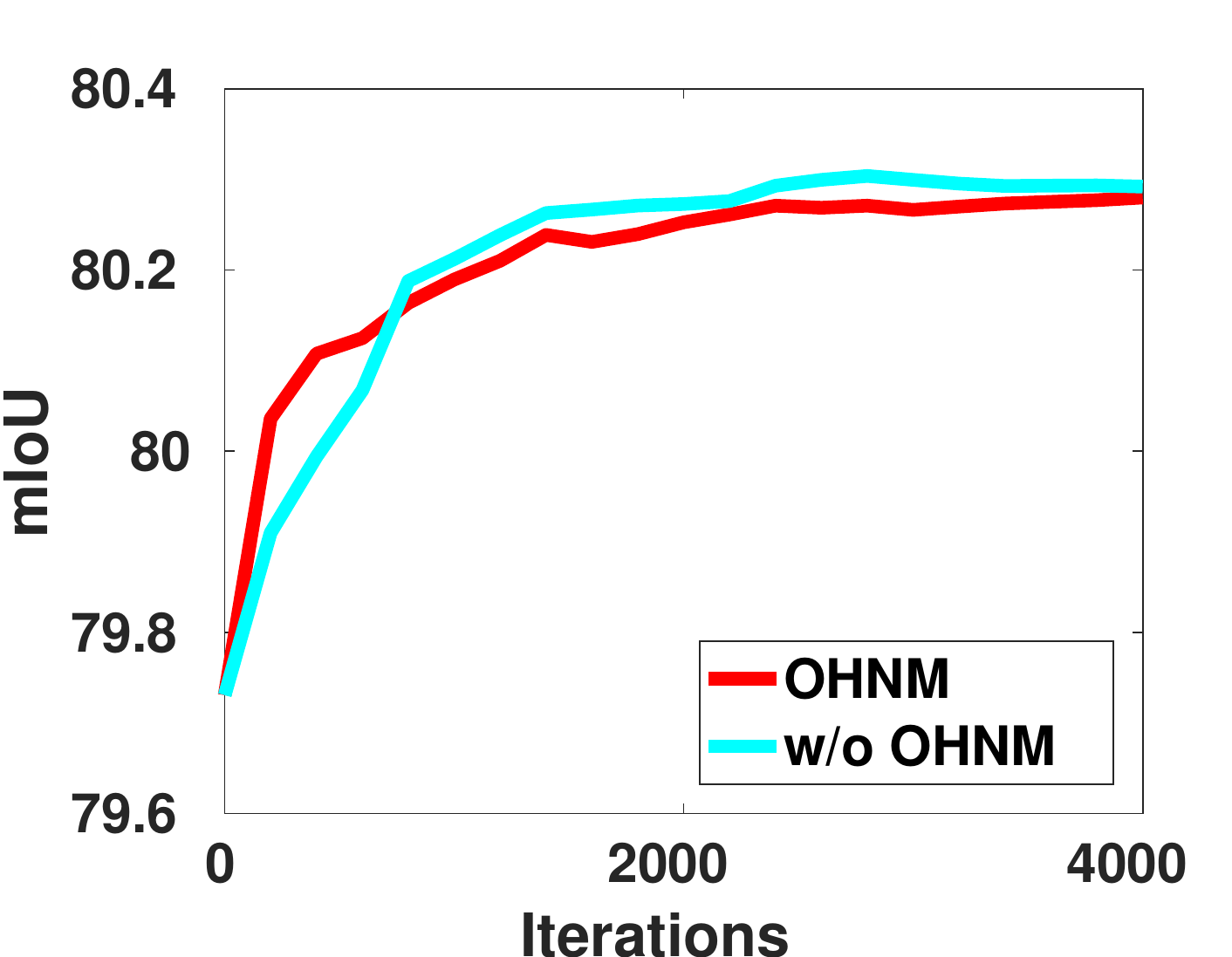}
   \end{center}
   \vspace{-0.4cm}
   \caption{ClassAcc and mIoU on \textit{Cityscapes} w.r.t. applying online hard negative mining (OHNM). }
   \vspace{-0.5cm}
\label{fig:pspnet_cityscapes_ohnm}
\end{figure}

\myparagraph{Effectiveness of online hard negative mining}
In Fig.~\ref{fig:pspnet_cityscapes_ohnm}, we compare the models with randomly sampled masks and mined hard negative during training as described in section~\ref{subsec:ohnm}. To begin with, we are able to observe that simply adding synthetic features is already helping to improve the performance, which demonstrates the effectiveness of our augmentation pipeline with synthetic features. Additionally, how to make full use of synthetic features to compensate for the data distribution for boosted performance is also interesting. Even though there may be better designs to incorporate synthetic features, the online hard negative mining is simple and effective, which shows the necessity of feature-level data augmentation.

\vspace{-0.1cm}
\section{Conclusion}
\vspace{-0.2cm}
\label{sec:conclusion}
Designing advanced network architectures to incorporate context more suitable or extract more representative features are important in semantic segmentation. We improve semantic segmentation from a different avenue according to utilizing data more effectively.
We propose synthesizing convolutional features for improving semantic segmentation.
Our pipeline is simple but surprisingly effective to reach stronger segmentation results.
A powerful architecture for feature synthesis is presented, enabling us to synthesize realistic features with rich details. Also, several techniques are presented to leverage synthetic features more effectively for the success of improved semantic segmentation.

{
\vspace{-0.3cm}
\bibliographystyle{splncs04}
\bibliography{egbib}
}
\end{document}